# UAV Path Planning Employing MPC-Reinforcement Learning Method Considering Collision Avoidance

Mahya Ramezani
*Centre for Security, Reliability, and Trust*
*University of Luxembourg*
Luxembourg, Luxembourg
mahya.ramezani@ext.uni.lu

Hamed Habibi
*Centre for Security, Reliability, and Trust*
*University of Luxembourg*
Luxembourg, Luxembourg
hamed.habibi@uni.lu

Jose Luis Sanchez-Lopez
*Centre for Security, Reliability, and Trust*
*University of Luxembourg*
Luxembourg, Luxembourg
jouseluis.sanchezlopez@uni.lu

Holger Voos
*Centre for Security, Reliability, and Trust*
*University of Luxembourg*
Luxembourg, Luxembourg
holger.voos@uni.lu

*Abstract*—In this paper, we tackle the problem of Unmanned Aerial (UAV) path planning in complex and uncertain environments by designing a Model Predictive Control (MPC), based on a Long-Short-Term Memory (LSTM) network integrated into the Deep Deterministic Policy Gradient algorithm. In the proposed solution, LSTM-MPC operates as a deterministic policy within the DDPG network, and it leverages a predicting pool to store predicted future states and actions for improved robustness and efficiency. The use of the predicting pool also enables the initialization of the critic network, leading to improved convergence speed and reduced failure rate compared to traditional reinforcement learning and deep reinforcement learning methods. The effectiveness of the proposed solution is evaluated by numerical simulations.

## I. Introduction

Nowadays, Unmanned Aerial Vehicles (UAVs) are being increasingly used in various industries, e.g., agriculture, mining, and construction, with different applications, such as monitoring, inspection, detection, reconnaissance, and mapping [1-3]. For this, UAV path planning plays a vital role in determining and tracking efficient and safe flight paths considering terrain, weather conditions, and dynamic obstacles.

Path planning, in the context of UAV navigation, refers to determining the best route for the UAV to follow while avoiding obstacles and reaching its destination [4, 5]. Early path planning methods were mainly graph-based, which could not handle dynamic obstacles [6, 7]. Subsequently, Artificial Potential Field (APF), Rapid exploration Random Tree (RRT), and D* algorithms were proposed as solutions for dynamic environments[8, 9]. However, APF can be stuck in local minima, RRT is unsuitable for continuous, highly dynamic environments, and the planned path is usually suboptimal and nonsmooth. Even though the D* algorithm effectively finds a path in a dynamic environment, convergence is still challenging in complex environments with many obstacles and dynamic obstacles.

To resolve these issues, Reinforcement Learning (RL) has emerged as a promising approach for UAV path planning. Salient features of RL, such as learning from experience and improving decision-making over time, make it a well-suited approach for UAV path planning in dynamic and uncertain environments[10]. RL can overcome traditional methods' limitations by allowing the UAV to adapt to time-varying conditions and obstacles. Additionally, the uncertainty and randomness of complex environments can be tackled by RL-based algorithms [11]. However, the poor performance of RL methods for handling high-dimensional systems is the main drawback that might make it impractical.

Deep Reinforcement Learning (DRL) is a recent Artificial intelligence (AI) technique that combines the strengths of Deep Learning and RL to solve decision-making problems for high-dimension systems. Therefore, DRL has been applied to UAV path planning in complex scenarios [12], resulting in more efficient and accurate solutions. For instance, in [13], the path smoothness, efficiency, and robustness were demonstrated. Furthermore, similar studies highlight the practical benefits of DRL for improving UAV path planning, addressing the challenges of dynamic obstacle avoidance and real-time decision-making in complex environments while handling the high-dimension issue [13-15].

Moreover, DRL methods require an approximation of the action-value function, often using deep neural networks (DNNs) [16]. However, it can be challenging to analyze the performance of a system controlled by an RL algorithm with an approximate policy represented by a DNN or another function approximation, which is especially important for systems with critical safety requirements. Furthermore, DNN-based RL has limitations regarding stability analysis, state/input constraints satisfaction, and meaningful weight initialization [17].

Recently, Model Predictive Control (MPC)-based RL approaches have been proposed to overcome these limitations, which suggests using MPC as the function approximation for the optimal policy in RL. MPC involves solving an optimal control problem at each time step to determine the best control policy [18]. As a result, MPC not only generates an optimal input and state sequence for the entire prediction horizon but also has the ability to anticipate future actions based on the predicted state [19]. MPC can handle complex and uncertain environments by providing a structured approach to control. On the other hand, the DRL component provides the ability to learn from experience and optimize control performance over time., It also can adapt to new situations and improve performance over time.

The main drawback of MPC is the high computational complexity, which can be tackled by using semi-accurate models of the controlled process to provide predictions [20]. It should be mentioned that the predictive model's accuracy directly impacts the MPC framework's performance. Utilizing a DNN to predict the model provides a more accurate and reliable representation of the system dynamics, thereby enhancing the effectiveness of the MPC framework [21]. The Long-Short-Term-Memory (LSTM) network, as a DNN, can extract sequential features and handle complex dependencies, making it ideal for predicting sequential processes [22].

Motivated by the considerations mentioned above, in this paper, we propose an LSTM-based-MPC system that operates as a deterministic policy within the Deep Deterministic Policy Gradient (DDPG) network, providing a robust and efficient solution for UAV path planning problem in unknown environments. Integrating the LSTM-MPC solution into the DDPG framework addresses the challenges posed by real-time path planning and results in improved convergence speed and reduced failure, offering a promising solution for path planning problems in complex and uncertain environments. The main contributions of this paper are as follows.

• Combining LSTM -MPC and DDPG algorithms to handle the high dimensionality of the state space, the uncertainty and variability of the obstacles' trajectories, and the real-time constraints of the system.

• Implementation of the LSTM-MPC-based model as an actor network in the DDPG framework

• Defining a predicting pool for predicting future state and corresponding actions

The rest of the paper is structured as follows. Section 2 outlines the path planning problem. Section 3 describes the proposed LSTM-MPC DDPD method. Section 4 presents the experiments carried out to assess the algorithm's performance. Concluding remarks are given in Section 5.

## II. PROBLEM FORMULATION AND PRELIMINARIES

This paper proposes a novel method for path planning for UAVs in unknown environments with static and dynamic obstacles. The problem is safe path planning in case of limited knowledge of the environment. Path planning aims to find the optimal path in considering obstacle avoidance. The UAV is equipped with a LIDAR for obstacle avoidance and a GPS for positioning.

### A. Problem definition

The environment is modeled as a Markov Decision Process. The UAV is equipped with a LIDAR for obstacle avoidance and a GPS for positioning. The GPS sensor provides precise navigation and localization capabilities, enabling the UAV to maintain its position and navigate to the target point with high accuracy. The LiDAR sensor, on the other hand, offers real-time 3D mapping and obstacle detection capabilities, enabling the UAV to perceive and avoid dynamic obstacles present in the field. The UAV dynamics are presented in [23], which is not repeated here for brevity.

Here, the safe UAV path planning problem in an unknown environment with static and dynamic obstacles is tackled with limited knowledge of the environment.

### B. The Markov Decision Process

The Markov Decision Process (MDP) is a well-established mathematical framework for modeling decision-making problems in the presence of uncertainty. MDP is defined by a tuple ⟨S, A, P, R, γ⟩, where S denotes a finite set of states, A denotes a finite set of actions, P represents the state transition probabilities as

$$p_{ss'}^a = P[S_{t+1} = s' \mid S_t = s, A_t = a] \quad (1)$$

R denotes the reward function as

$$R_s^a = E[R_t \mid S_t = s, A_t = a] \quad (2)$$

and $\gamma$ is the discount factor, which lies in the interval [0, 1] and with a stage cost $l(s,a)$.

Consider deterministic policy delivery of the control input $a = \pi(s)$, resulting in state distribution $\tau^\pi$. RL's goal is to find the optimal policy $\pi^*$ by solving the minimization problem $\pi^* \coloneqq \arg\min J(\pi) = \mathbb{E}_{\tau^\pi}[\sum_{k=0}^{\infty} \gamma^k l(s_k, \pi(s_k))]$.

In RL, important quantities are the action-value function $Q^*(s,a)$ and value function $V^*(s)$ associated with the optimal policy $\pi^*(s)$, which are defined by the Bellman equations

$$Q^*(s,a) = l(s,a) + \gamma \mathbb{E}[V^*(s_{t+1}) \mid s, a] \quad (3)$$
$$V^*(s) = Q^*(s, \pi^*(s)) = \min_a Q^*(s,a) \quad (4)$$

### C. MDP formulation

#### 1) State

The proper action is chosen based on the current state of the UAV. The obstacle is cylindrical shape and unknown in the environment. Moreover, the shape of the environment is a rectangle. The state of the UAV is represented by a state vector consisting of the position of the UAV in the world frame, represented by $(x, y, z)$, and the velocity of the UAV along the X, Y, and Z axes, represented by $(v_x, v_y, v_z)$.

For ease of navigation towards the target, the absolute position of the UAV is transformed into its relative position with respect to the target. The environment information is gathered using Lidar distance sensors, with a scan angle range of $\pi$ and an angle of $\pi/6$ between each pair of rays. The environment observations are divided into 7 sensors in the horizontal plane and 7 in the vertical plane. The state represented:

$$S = [x, y, z, v_x, v_y, v_z, \phi, \theta, \psi, x_t, y_t, z_t, d0, \ldots, d_N] \quad (5)$$

To simplify the description of the UAV's position and motion, a three-degree-of-freedom kinematic model is adopted in this study. The assumption is that the UAV maintains a constant horizontal altitude, thus confining its movement to the x-y plane. By disregarding the momentum effects during flight and assuming a constant velocity v, the position, and motion of the UAV are represented by the vector $\zeta = [x, y, \psi]$.

Five variables represent the state of the UAV in 2D $(x, y)$ for its position in the world frame $(v_x, v_y)$ for its velocity along the X and Y axes, and $\psi$ for the angle between its first-perspective direction and the line connecting the UAV to the target.

The relative position between the UAV and the target can be transformed to make it easier to reach the target. The environment information is obtained using laser distance sensors, which have a scan angle range of $\pi$ and an angle between each two laser rays of $\pi/6$. The observations are 7 sensors in the horizontal plane. And the state represented for UAV in a simplified environment is as follows

$$S = [x_t - x, y_t - y, v_x, v_y, v, \psi, d0, \ldots, d_N] \quad (6)$$

*2) Action*

the control of the UAV is simplified to only two commands - speed and yaw angle. The control vector of the UAV is represented as $a = [a_v, a_\psi]$ where $a_v$ is the current speed ratio to the maximum speed, with a range of values between -1 and 1. $a_\psi$ is a steering signal that determines the desired yaw angle, with a range of values between -1 and 1.

*3) Reward function*

The reward function for the UAV in the given scenario is a combination of four parts. It aims to guide the UAV to reach the target area while ensuring its safety, defined as follows.

Distance Reward: The negative value of the distance between the UAV and the target point is used as a penalty to encourage the UAV to reach the target area. The UAV will receive a positive reward if it reaches the target area. The relative distances represent the difference between the present location and the destination. We assume the difference between the UAV and its target in the previous and current stages, respectively.

$d_{dis} = d_{curent} - d_{prevous}$

$$r_1 = \begin{cases} -0.1 & d_{dis} > 0 \\ 1 & d_{dis} < 0 \end{cases} \quad (7)$$

*Step Penalty*: To make the UAV reach the target area as soon as possible, a penalty of $r_2 = -0.01$ is given to the UAV at each step.

*Orientation Penalty*: the negative value of the angle between the UAV's first-perspective direction and the connection line of the UAV and the target is taken as a penalty to help the UAV approach the target in the direction towards the target. It is considered $r_3 = -0.01$.

*Obstacles Penalty*: To keep the UAV away from obstacles, first should define a safe distance to prevent collision with the obstacles. When the distance between the UAV and the nearest obstacle is less than the safe distance, the UAV receives a penalty for the distance. If the UAV collides with obstacles, it receives another penalty. Otherwise, it does not receive any punishments if the distance exceeds the safe distance.

$$r_4 = \begin{cases} -0.1 & d_{safe} > 0 \\ -1 & collision \\ 0 & else \end{cases} \quad (8)$$

Therefore, the reward received by the UAV at the current moment can be expressed as follows:

$$R = w_1 r_1 + w_2 r_2 + w_3 r_3 + w_4 r_4 \quad (9)$$

To mention more attention on collision avoidance to reduce risk and have safe path planning and the optimal path, we chose the weight of the $w_1$ and $w_4$ more than $w_2$ and $w_3$

### III. METHODOLOGY

#### A. Deep Deterministic Policy Gradients

Deep Deterministic Policy Gradients (DDPG) is a popular DRL algorithm that combines elements of both value-based and policy-based RL methods. It operates within the framework of an actor-critic architecture where the actor network is responsible for selecting actions based on the current state of the environment. The critic network is used to judge the value of the actions taken by the actor. The two networks are trained simultaneously using the same set of experiences collected by the agent during its interactions with the environment. In order to overcome the problems of correlation between samples in reinforcement learning, the DDPG algorithm employs a replay buffer that stores experience and random samples from this buffer when updating the networks. DDPG is derived from the deterministic policy gradient theorem for Markov Decision Processes (MDPs) with continuous action spaces, trains the networks using stochastic gradient descent with mini-batches, and updates the target networks using a soft update algorithm. It is a visual illustration of the actor-critic architecture, including the Q network and μ network, and their interactions with the environment. The use of a target network and the replay buffer are also depicted in Figure 1, highlighting their role in improving the stability and sample efficiency of the algorithm.

#### B. The Long Short-Term Memory Model Predictive Control

*1) LSTM*

LSTM is a Recurrent Neural Network (RNN) type that features external recurrences from the outputs to the hidden layer's inputs and internal recurrence between LSTM cells. In each LSTM cell, a set of gating units regulates the information flow, enabling the network to remember or forget information based on a sequence of inputs [24]. The LSTM network consists of an input layer, multiple hidden layers, and an output layer. Its uniqueness lies in its memory cells in the hidden layers, which allow data to be learned by maintaining or adjusting the memory cell's state. The proposed LSTM model was built using the deep learning toolbox in MATLAB. To optimize the LSTM structure, learning parameters are specified and summarized in Table 1. The set values for these parameters were chosen based on recommendations from MATLAB and previous research.

Table 1. The LSTM hyperparameters.

| Parameter | Value |
|---|---|
| Optimization algorithm | Adam |
| Initial learn rate | 0.01 |
| Hidden layer | 3 |
| Hidden unit | 200(3) |
| Max epochs | 100 |

*2) The Long Short-Term Memory Model Predictive Control*

The Long Short-Term Memory Model Predictive Control (LSTM-MPC) is a novel control algorithm that combines the strengths of Model Predictive Control (MPC) and LSTM networks. The LSTM network models the state dynamics of a system, while MPC is used to generate control actions based on state predictions. LSTM-MPC aims to find an optimal control sequence, $u^*$, that minimizes a cost function, J, over a set of predicted states and control actions in real-time environments.

The LSTM network predicts state dynamics based on the current state and control actions. Given the initial states, $s_0$, and control actions, u, the LSTM network predicts the next state, $s_{\{t+1\}}$.

MPC generates control actions, u, based on predicted states, $s$. The control actions are computed by minimizing the cost function, $J$, over predicted states and control actions using a quadratic program (QP). The predicted states and control actions are inputs to the optimization problem, and the optimal control actions are the outputs.

The algorithm offers several advantages, including improved state prediction accuracy, handling unknown state dynamics, and incorporating constraints on control actions and states. On the other hand, it can face Several challenges that can arise when using LSTM-MPC for path planning with collision avoidance and considering unknown dynamic obstacles. Some of these challenges include LSTM-MPC has limitations, including sensor noise and measurement errors, model uncertainty, computational complexity, nonlinearity and non-convexity, and data efficiency. The optimized control action may also be insufficient due to model mismatch, nonlinear dynamics, unmodeled disturbances, limited prediction horizon, constraint violations, and real-time requirements.

In this paper, The LSTM network and MPC algorithm are trained together in an iterative process, with the LSTM network improving its model dynamics based on the feedback from the MPC algorithm and the MPC algorithm using the improved LSTM model to make better control decisions. After each iteration of the MPC algorithm, the predicted following states and the following actual states can be compared, and the difference between them can be used to update the parameters of the LSTM network using an optimization algorithm such as gradient descent. The training goal is to minimize the prediction error between the actual and predicted following states, thus improving the accuracy of the LSTM model dynamics.

*C. LSTM-MPC-DDPG*

The proposed algorithm for RL in UAV path planning is based on integrating DDPG and LSTM-MPC algorithms. The integration of these two algorithms leverages their respective strengths to provide a robust solution for real-time constraints and unknown environments.

The integration of MPC and LSTM allows the algorithm to make informed decisions in unknown environments by combining the strengths of both algorithms. The MPC component generates training data for the LSTM network by repeatedly applying the policy to the system and collecting the resulting states and actions. The LSTM network is trained using this data to predict future states and actions. The MPC component can then use the predictions made by the LSTM network to make more informed decisions about which actions to take in the future.

The LSTM-MPC component is utilized as a deterministic policy, where the LSTM network is pre-trained using limited knowledge of the environment map. The future state action predictions generated by the LSTM-MPC are stored in a "predicting pool." The predicted data in predicting pool will be used to improve the training of the critic network. One approach is to use the predicted data as a warm start for the critic network by initializing the network parameters using the estimated Q-values for each future state-action pair. This can help to speed up the convergence of the network.

Additionally, the estimated Q-values can be used as target values during the backpropagation step, allowing the network to learn to predict better the quality of the actions taken by the policy and make more informed decisions about which actions to take the future. This approach can lead to improved accuracy in the Q-value estimates and more efficient network convergence. These estimated Q-values can be used as the target values for the critic network during the backpropagation step, where the loss is calculated as the mean-squared error between the estimated Q-values and the actual Q-values.

Using the "predicting pool" for initialization combined with warm starting further enhances the algorithm's efficiency. Using a "predicting pool" in the DDPG algorithm provides several benefits that improve the performance and efficiency of the algorithm.

- Speed up the learning process: By providing an initial estimate of the expected return, the learning process can be initiated from a better starting point. This reduces the number of iterations required for convergence to an optimal solution, thus speeding up the learning process.
- Improved accuracy of predictions: The data collected from the "predicting pool" can be utilized to refine the accuracy of predictions made by the LSTM-MPC (Long Short-Term Memory Model Predictive Control), which leads to improved performance of the DDPG algorithm.
- Better utilization of data: The "predicting pool" serves as a repository for the data generated by the LSTM-MPC predictions. This facilitates the better utilization of the data and enables easy incorporation into the learning process.
- Increased stability: By utilizing the data from the "predicting pool" as supplementary information and

combining it with the observations from the environment to estimate the TD (Temporal difference) error, the algorithm's stability can be improved. The weight assigned to the predicted data can be adjusted dynamically based on its reliability, and the observed data can serve as a backup to ensure stability and accuracy.

In this work, the LSTM-MPC algorithm is utilized as a function approximator in the actor network of the DDPG framework for UAV path planning in the x-y plane with a constant altitude. The cost function for UAV path planning is formulated as a combination of several terms: the deviation of the UAV's position from the target point, a collision avoidance term based on the lidar data, and the magnitude of the control inputs required to execute the path. The optimization problem, defined as the minimization of this cost function, is solved at each time step by the MPC algorithm to determine the control inputs, which are treated as the deterministic policy for the actor-network in the DDPG framework. The LSTM network and MPC parameters are updated over time as the system learns from its interactions with the environment.

The cost function in the MPC algorithm for a quadrotor in an x-y plane at a constant altitude can be formulated as a trade-off between following the optimal path and avoiding collisions. One possible formulation of the cost function could be

$$J_{MPC}(x,u) = w_1 \left\| x_d(t+j) - x_p(t+j) \right\|^2 + w_2 \|\Delta u(t+j)\|^2 + w_3 C_{collision}(x) \quad (10)$$

where $x_p(t+j)$ is the current state of the system (position and velocity) based on the LSTM network at the $(t+j)^{th}$ period based on the available measurements at the $t^{th}$ sampling period, $u$ is the control input, $x_d$ is the desired state, $w_1$, $w_2$, and $w_3$ are weighting factors that balance the importance of following the desired state, minimizing control effort, and avoiding collisions, respectively. More details regarding the LSTM_MPC formulation can be found in [25].

The collision cost in the context of a quadrotor in the x-y plane can be formulated as a measure of the distance between the quadrotor and any potential obstacles in the environment. the cost can be defined as the negative exponential of the Euclidean distance between the quadrotor's current position and the closest obstacle. Formally, the collision cost can be defined as:

$$C_{collision} = -e^{(-d(q,obs))} \quad (11)$$

Where $d(q,obs)$ is the Euclidean distance between the quadrotor's position q and the closest obstacle, the negative exponential ensures that the cost increases as the quadrotor approaches an obstacle and approaches infinity as the distance decreases to zero, encouraging the quadrotor to avoid collisions.

This cost function can be added to the overall cost function used in the MPC optimization problem, which may also include other objectives such as reaching a target position, tracking a desired trajectory, or minimizing control inputs. The optimization problem is then solved at each time step to determine the control inputs for the quadrotor, which can be treated as the deterministic policy for the actor network in a DDPG framework.

In order to evaluate the reward for each set of states, action, and future state in the MPC-based deterministic policy, it is necessary to incorporate the reward function into the MPC optimization process. The reward function assigns a numerical value to the quality of a given state-action-future state combination. It serves as a means of reinforcing desirable actions and discouraging undesirable ones. To this end, the MPC optimization problem can be formulated as follows:

$$min J_t = J_{MPC}(x_t, u_t) + \sum w_i R(s_i, a_i) \quad (12)$$

where $J_t$ is the objective function, $w_i$ is the weight that controls the relative importance of each term, $R(s_i, a_i)$ is the reward function, and the constraints represent operational limitations and requirements for the UAV. Including the reward function in the MPC optimization makes it possible to evaluate the reward for each set of states, actions, and future states. These rewards can then be stored in the predicting pool, providing valuable information that can be used to improve the accuracy of the LSTM network and the performance of the MPC algorithm and the Actor-Critic Network. Overall, incorporating the reward signal into the MPC optimization process guides the UAV toward making better decisions in unknown environments.

*D. LSTM-MPC as function approximation*

The LSTM-MPC actor network is trained to maximize the expected cumulative reward through gradient descent. The gradient of the expected cumulative reward with respect to the network parameters (denoted as θ) is computed using the chain rule of differentiation as

$$\nabla \theta J(\theta) = \nabla \theta \mathbb{E}[R|\theta] \quad (13)$$

where $J(\theta)$ is the expected cumulative reward, $\mathbb{E}[R|\theta]$ is the expected cumulative reward given the current policy (parameterized by $\theta$), and $\nabla \theta$ denotes the gradient with respect to the parameters $\theta$. The gradient is estimated using the experience gathered from the environment. The LSTM-MPC actor network is updated after each episode or after a batch of episodes to maximize the expected cumulative reward.

The parameters of the LSTM network, theta, are trained using gradient descent to minimize the expected cost over a sequence of states and control inputs. In each iteration of the MPC algorithm, the LSTM network uses the current state, $x_t$, and the past control inputs, $u_{t-1}, \ldots, u_{t-k}$, to predict the control input for the next time step, $u_t$. The predicted control input, $u_t$, is then used in the optimization problem to determine the optimal control input, $u^*$. The parameters of the LSTM network are updated using gradient descent as follows

$$\theta = \theta - \alpha \nabla (J(x, u^*), \theta) \quad (14)$$

where $\alpha$ is the learning rate.

By iterating this process, the parameters of the LSTM network converge to the optimal policy, $\pi^*$, that maps states to control inputs to minimize the expected cost. The policy can be used to control the system in real time. A further detailed formulation can be found in [26].

*1) Steps of the proposed algorithm*
**Step 1**: Training of LSTM Network:
- Initialize the LSTM network using limited knowledge of the environment.

- The LSTM network is initialized with parameters $\theta_0$. The network's hidden state, $h_0$, is initialized based on the initial conditions of the system.

**Step 2**: Improvement of LSTM with Real-time Sensor Data:
- Use real-time data from sensors to update the network's predictions continually.
- This allows the network to adapt to environmental changes and continuously improve its predictions.
- Use backpropagation and gradient descent to update the network parameters, w, to minimize the prediction error.

**Step 3**: Use of LSTM to Predict the Environment and Control Action:
- The LSTM network is used to predict the environment, while the Model Predictive Control (MPC) is used to determine the optimal control action. The LSTM considers the MPC data and past data to predict the next state.
- The control actions are generated by minimizing the cost function, J, over a set of predicted states and control actions, as described in the previous sections.
- LSTM uses MPC data to train and improve the MPC network

**Step 4**: LSTM-MPC as Deterministic Policy Actor-Network:
- Use the LSTM-MPC network to determine the optimal control actions for a given state.
- Parametrized LSTM based on the $\theta$ to use gradient descent for updating the policy

**Step 5**: Predicting Pool:
- Use the LSTM-MPC network to predict future states and corresponding control actions in a set MPC horizon.
- Evaluate the rewards for each control action
- Store information in a new experience pool called the "predicting pool."

**Step 6**: Collection of Experiences from the Environment:
- Interact with the environment by choosing actions and collecting information about the current state, action, reward, and next state.
- Store this information in a separate experience pool.

**Step 7**: Training of Critic Network:
- Use a critic network to estimate the value function for each experience in the batch from the "experience pool."
- Using the predicting states from the LSTM network for more sample
- Calculate the target value for each experience, which is the reward plus the estimated value of the next state.
- Train the critic network to predict the target value by minimizing the mean squared error between the prediction and the target.
- The "predicting pool" is used to initialize the value function approximator, which can speed up the training process and reduce the number of iterations required to converge. (Warm start)

**Step 8**: Improvement of LSTM-MPC:
- Use the data from the "predicting experience pool" to warm-start the LSTM-MPC, improving the accuracy of its predictions based on their reliability.
- To improve the real-time performance of the MPC, efficient optimization algorithms and approximations can be employed to reduce computational complexity.
- These algorithms can significantly improve the speed and efficiency of the MPC, allowing it to make predictions and choose actions in real time.

**Step 9**: Repeating Steps 2-8:
- Continuously repeat Steps 2 through 8 to improve the MPC's performance.

The algorithm structure block diagram is shown in Figure 1.

IV. SIMULATION RESULTS AND SETTING

*A. Experiment enviorment*

In this scenario, the environment is a simplified 3D space with dimensions of 200m x 200m x 50m. The target and initial points of the UAV are randomly initialized in one of the corners of the environment. Cylindrical Obstacles of different sizes are randomly generated within the rest area. In order to navigate this environment, the UAV uses its sensor data to avoid obstacles and reach its target point. The environment was chosen to be simplified to make the task of path planning and obstacle avoidance more manageable. The safe distance to the obstacle is 1.5m, and the max sensor length is 20m.

*B. Training*

The proposed algorithm in this paper operates within the MATLAB environment and utilizes a learning rate of 0.001 for its optimizer. The maximum number of iterations is set to 5000 to ensure network convergence, and the maximum number of steps per iteration is set to 500. the experience pool has a maximum capacity of $10^5$. The episode ends if the UAV reaches the target area, collides with the obstacles, and the number of training steps reaches the maximum. The soft update factor $\tau$ is 0.001, and the exploration noise is set to 0.2. the MPC horizon is set as 5 with time step 0.5 and minimum batch size $N = 256$.

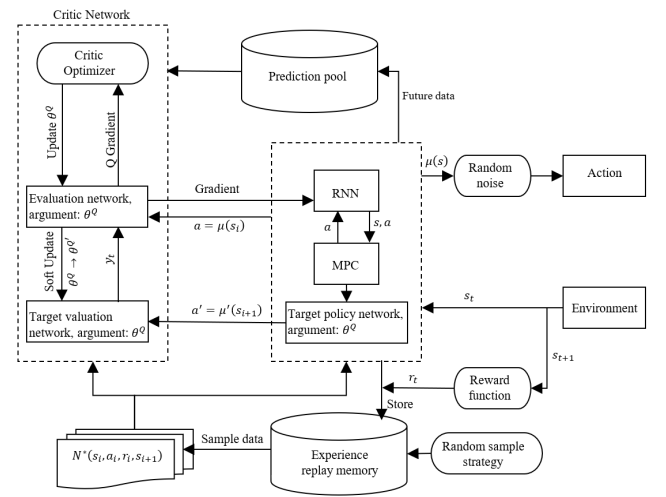

Figure 1. The diagram of the proposed algorithm.

In the training environment, there are five cylindrical obstacles with a size of 5 x 50 and five cylindrical obstacles with a size of 10 x 50. Figure *2* and Figure *3* show the training data's success ratio and cumulative reward function. In the initial training stage, the UAV operates randomly due to the lack of accurate prediction, resulting in a low average reward. However, the MPC-based deterministic policy shows a higher average reward than the other two methods at the beginning and the end of episodes, the proposed algorithm also converges faster. The success rate represents the percentage of successful target acquisitions.

*C. Results*

After 5000 episodes of training in the training environment for DDPG, TD3, and the proposed algorithm, they were tested in dynamic and static environments. The proposed algorithm was first evaluated in a static eenvironment, where obstacles were randomly placed in each episode to assess the algorithm's ability to generalize. The environment uses two different settings; E1 and E2. E1 consists of five cylindrical obstacles with dimensions of 20x50 and five with dimensions of 15x50. E2 consists of ten obstacles with dimensions of 20x50 and ten with dimensions of 5x50.

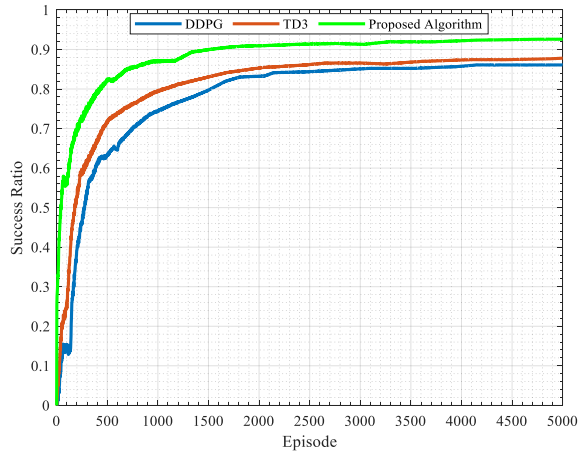

Figure 2. The success ratio of reaching the destination in the training phase.

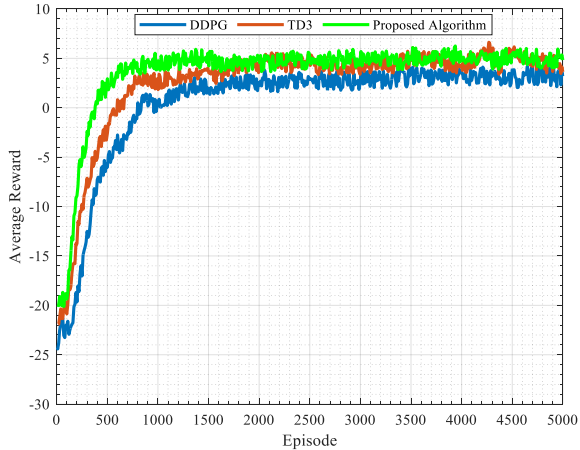

Figure 3. The comparison of the average reward of the algorithms in the training phase.

The results of the proposed algorithm were compared to those of the DDPG and TD3 algorithms after 3000 episodes. As shown in Table 2, the proposed algorithm demonstrated the highest success rate in all three settings. The advantage of the proposed algorithm became increasingly apparent in more complex environments with a higher number of obstacles. In the more complex environment, E2, the success rate of the proposed algorithm decreased less than two others.

Furthermore, The performance of the proposed algorithm, DDPG, and TD3 algorithms was evaluated in terms of cumulative step convergence. Results indicated that the proposed algorithm demonstrated faster convergence speed when compared to the TD3 and DDPG. Upon testing the algorithms for 3000 episodes in the same environment, These findings suggest that the proposed LSTM_MPC_DDPG algorithm exhibits superior convergence and performance compared to the TD3 and DDPG algorithms. Figure 4 and Figure 5 show the paths generated in environments E1 and E2 based on the proposed algorithm, respectively, where the black dashed line shows the safe zones around obstacles. As seen, the UAV reaches the target area without collision with obstacles using a short path.

Next, algorithms were tested in a dynamic environment. In this scenario, obstacles move along the X-axis at different speeds, which can test whether the UAV can make real-time decisions in dynamic environments. Obstacles move at the speed of v in the negative direction of the X-axis. DDPG, TD3, and the proposed algorithm are tested for 3000 episodes in 2 different speeds of obstacle. The results in different dynamic environments are illustrated in Table 2. The success rates of the three algorithms decrease as the speed of obstacles increases. The success rate of the proposed algorithm is the lowest.

Moreover, the success rate of our algorithm is much higher than that of TD3. It is because the design of the predicting pool network and MPC_LSTM as an actor-network structure works. UAVs can perceive the changes in the surrounding environment on the horizon by adding environmental information changes to predicting pool inputs so that obstacles can be avoided in time. The results show that the proposed algorithm has strong adaptability to dynamic environments.

Table 2. Tests results under static environments and the impact of different speeds of a dynamic obstacle in a dynamic environment

|  |  | E1 | E2 | V=10 m/s | V=15 m/s |
|---|---|---|---|---|---|
| DDPG | SR (%) | 92.25 | 80.5 | 85.2 | 78.15 |
|  | CR (%) | 7.05 | 10.1 | 14.3 | 21.75 |
|  | LR (%) | 0.7 | 9.4 | 0.5 | 0.1 |
|  | AR | 3.42 | 2.3 | 3.1 | 1.2 |
| TD3 | SR (%) | 94.1 | 85.7 | 90.25 | 85.25 |
|  | CR (%) | 4.8 | 8.2 | 8.5 | 14.15 |
|  | LR (%) | 1.1 | 6.1 | 1.5 | 0.6 |
|  | AR | 4.23 | 2.71 | 4.01 | 1.7 |
| Proposed Algorithm | SR (%) | 95.1 | 93.4 | 93.35 | 86.26 |
|  | CR (%) | 4.9 | 6.3 | 6.65 | 13.74 |
|  | LR (%) | 0 | 0 | 0 | 0 |
|  | AR | 5.08 | 3.02 | 4.25 | 2.56 |

**Success rate (SR):** The percentage of successful target acquisitions.
**Collision rate (CR):** The percentage of collisions with obstacles.
**Loss rate (LR):** The percentage of getting lost in the map (without collision).
**Average reward (AR):** The overall performance quality.

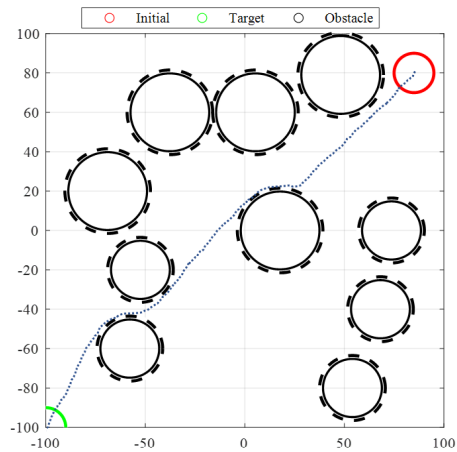

Figure 4. The generated path by the proposed algorithm in the E1 environment.

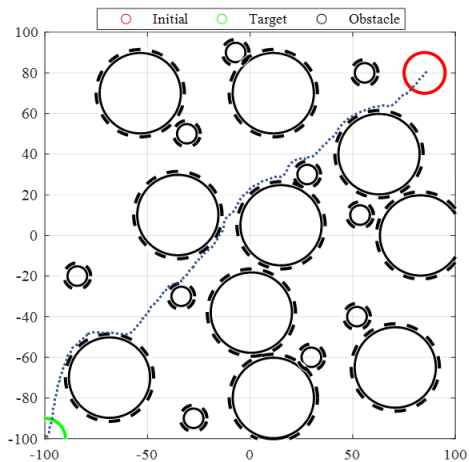

Figure 5. The generated path by the proposed algorithm in the E2 environment.

## V. Conclusion

This paper presents a novel method that leverages Deep Reinforcement Learning to facilitate autonomous path planning for unmanned aerial vehicles in complex environments with multiple static and dynamic obstacles. The proposed solution employs a deterministic policy based on the LSTM-MPC architecture to enhance the performance of the DDPG algorithm. The LSTM-MPC can predict environmental features from observations, thereby addressing the unpredictability and variability in these scenarios. The method incorporates a "predicting pool" that stores the set of state-action-reward data, which accelerates the performance of the DDPG algorithm. The efficacy of the proposed algorithm was evaluated in simplified 3D-simulation environments and compared with the DDPG and TD3 algorithms. The results demonstrated that the proposed solution could effectively train UAVs for path planning in complex environments with real-time dynamic obstacle avoidance while reaching the target area promptly and safely with higher performance indices than DDPG and TD3 algorithms.

Acknowledgments

This research was partially supported by the European Union's Horizon 2020 project Secure and Safe Multi-Robot Systems (SESAME) under grant agreement no. 101017258. For the purpose of open access, the author has applied a Creative Commons Attribution 4.0 International (CC BY 4.0) license to any Author Accepted Manuscript version arising from this submission.

References

[1] R. Raj, S. Kar, R. Nandan, and A. Jagarlapudi, "Precision agriculture and unmanned aerial Vehicles (UAVs)," *Unmanned Aerial Vehicle: Applications in Agriculture and Environment,* pp. 7-23, 2020.

[2] S. A. H. Mohsan, M. A. Khan, F. Noor, I. Ullah, and M. H. Alsharif, "Towards the unmanned aerial vehicles (UAVs): A comprehensive review," *Drones,* vol. 6, no. 6, p. 147, 2022.

[3] S. Park and Y. Choi, "Applications of unmanned aerial vehicles in mining from exploration to reclamation: A review," *Minerals,* vol. 10, no. 8, p. 663, 2020.

[4] D. Wang, "Indoor mobile-robot path planning based on an improved A* algorithm," *Journal of Tsinghua University Science and Technology,* vol. 52, no. 8, pp. 1085-1089, 2012.

[5] L. Li, T. Ye, M. Tan, and X.-j. Chen, "Present state and future development of mobile robot technology research," *Robot,* vol. 24, no. 5, pp. 475-480, 2002.

[6] G. Hoffmann, S. Waslander, and C. Tomlin, "Quadrotor helicopter trajectory tracking control," in *AIAA guidance, navigation and control conference and exhibit*, 2008, p. 7410.

[7] R. Bohlin and L. E. Kavraki, "Path planning using lazy PRM," in *Proceedings 2000 ICRA. Millennium conference. IEEE international conference on robotics and automation. Symposia proceedings (Cat. No. 00CH37065)*, 2000, vol. 1: IEEE, pp. 521-528.

[8] Y. K. Hwang and N. Ahuja, "A potential field approach to path planning," *IEEE transactions on robotics and automation,* vol. 8, no. 1, pp. 23-32, 1992.

[9] J. Bruce and M. Veloso, "Real-time randomized path planning for robot navigation," in *IEEE/RSJ international conference on intelligent robots and systems*, 2002, vol. 3: IEEE, pp. 2383-2388.

[10] G.-T. Tu and J.-G. Juang, "UAV Path Planning and Obstacle Avoidance Based on Reinforcement Learning in 3D Environments," in *Actuators*, 2023, vol. 12, no. 2: MDPI, p. 57.

[11] B. G. Maciel-Pearson, L. Marchegiani, S. Akcay, A. Atapour-Abarghouei, J. Garforth, and T. P. Breckon, "Online deep reinforcement learning for autonomous UAV navigation and exploration of outdoor environments," *arXiv preprint arXiv:1912.05684,* 2019.

[12] S. Faryadi and J. Mohammadpour Velni, "A reinforcement learning‐based approach for modeling and coverage of an unknown field using a team of autonomous ground vehicles," *International Journal of Intelligent Systems,* vol. 36, no. 2, pp. 1069-1084, 2021.

[13] S. Zhang, Y. Li, and Q. Dong, "Autonomous navigation of UAV in multi-obstacle environments based on a Deep Reinforcement Learning approach," *Applied Soft Computing,* vol. 115, p. 108194, 2022.

[14] B. Xin and C. He, "DRL-Based Improvement for Autonomous UAV Motion Path Planning in Unknown Environments," in *2022 7th International Conference on Control and Robotics Engineering (ICCRE)*, 2022: IEEE, pp. 102-105.

[15] Y. Li and A. H. Aghvami, "Intelligent UAV Navigation: A DRL-QiER Solution," in *ICC 2022-IEEE International Conference on Communications*, 2022: IEEE, pp. 419-424.

[16] S. Gros and M. Zanon, "Reinforcement learning based on mpc and the stochastic policy gradient method," in *2021 American Control Conference (ACC)*, 2021: IEEE, pp. 1947-1952.

[17] S. Gros and M. Zanon, "Data-driven economic NMPC using reinforcement learning," *IEEE Transactions on Automatic Control,* vol. 65, no. 2, pp. 636-648, 2019.

[18] M. L. Darby and M. Nikolaou, "MPC: Current practice and challenges," *Control Engineering Practice,* vol. 20, no. 4, pp. 328-342, 2012.

[19] A. B. Kordabad, W. Cai, and S. Gros, "MPC-based reinforcement learning for economic problems with application to battery storage," in *2021 European Control Conference (ECC)*, 2021: IEEE, pp. 2573-2578.

[20] Z. Zhang, D. Zhang, and R. C. Qiu, "Deep reinforcement learning for power system applications: An overview," *CSEE Journal of Power and Energy Systems,* vol. 6, no. 1, pp. 213-225, 2019.
[21] H. Moumouh, N. Langlois, and M. Haddad, "A Novel Tuning approach for MPC parameters based on Artificial Neural Network," in *2019 IEEE 15th International Conference on Control and Automation (ICCA)*, 2019: IEEE, pp. 1638-1643.
[22] Y. Jiao, Z. Wang, and Y. Zhang, "Prediction of air quality index based on LSTM," in *2019 IEEE 8th Joint International Information Technology and Artificial Intelligence Conference (ITAIC)*, 2019: IEEE, pp. 17-20.
[23] B. Jiang, B. Li, W. Zhou, L.-Y. Lo, C.-K. Chen, and C.-Y. Wen, "Neural network based model predictive control for a quadrotor UAV," *Aerospace,* vol. 9, no. 8, p. 460, 2022.
[24] A. Sherstinsky, "Fundamentals of recurrent neural network (RNN) and long short-term memory (LSTM) network," *Physica D: Nonlinear Phenomena,* vol. 404, p. 132306, 2020.
[25] J. Yan, P. DiMeo, L. Sun, and X. Du, "LSTM-Based Model Predictive Control of Piezoelectric Motion Stages for High-speed Autofocus," *IEEE Transactions on Industrial Electronics,* 2022.
[26] W. Cai, A. B. Kordabad, H. N. Esfahani, A. M. Lekkas, and S. Gros, "MPC-based reinforcement learning for a simplified freight mission of autonomous surface vehicles," in *2021 60th IEEE Conference on Decision and Control (CDC)*, 2021: IEEE, pp. 2990-2995.